\documentclass[preprint,12pt]{elsarticle}

\usepackage{amssymb}
\usepackage{amsmath}
\usepackage{booktabs} 
\usepackage{multirow}
\usepackage{cleveref}
\usepackage{xcolor}
\usepackage{tabularx}
\usepackage{float}
\usepackage{url}

\journal{Nuclear Physics B}

\begin{document}

\begin{frontmatter}



\title{Lightweight Chunk Selection for Mobile Retrieval-Augmented Generation}


% \author{Yidan Shen, Yu Wen, Aamir Bader Shah, Xin Fu, Renjie Hu} 
\author[label1]{Sicong Chang\fnref{eq}} 
\author[label1]{Yidan Shen\fnref{eq}}
\author[label3]{Wen Yu}
\author[label1]{Jiefu Chen}
\author[label1]{Xin Fu}
\author[label3]{Renjie Hu\corref{cor1}}
\cortext[cor1]{Corresponding author.}
\ead{rhu7@central.uh.edu}
\fntext[eq]{Equal contribution.}
%% Author name

\affiliation[label1]{organization={Department of Electrical and Computer Engineering, University of Houston},%Department and Organization
            %addressline={4226 Martin Luther King Blvd}, 
            city={Houston},
            % postcode={77204}, 
            state={TX},
            country={USA}}

\affiliation[label3]{organization={Department of Information Science Technology, University of Houston},%Department and Organization
            %addressline={4226 Martin Luther King Blvd}, 
            city={Houston},
            % postcode={77204}, 
            state={TX},
            country={USA}}

%\cortext[cor1]{Corresponding author. \\ Email address: yshen28@cougarnet.uh.edu (Y. Shen).}

%% Abstract
\begin{abstract}
Retrieval-augmented generation (RAG) improves the factual grounding of large language models (LLMs) by incorporating external knowledge, but deploying RAG on mobile and edge devices remains challenging because retrieved context increases computation and memory. A direct way to reduce this cost is to retain only one retrieved chunk before generation, but the top-ranked retrieved chunk is not always the most evidence-supporting one, since retrieval similarity does not necessarily imply evidential sufficiency. Existing context-reduction methods, including prompt or token-level compression, and cross-encoder reranking, can improve context quality, but often require additional LLMs or compressors that are costly under a strict mobile budget. In this paper, we study lightweight RAG chunk selection as an evidence-alignment problem. Our selector combines three complementary feature sources: question hidden states that represent LLM-side query intent, MoE routing-derived expert signals that capture the generator's internal routing structure, and retrieved chunk embeddings that preserve candidate-side evidence geometry. A compact multilayer perceptron maps these features to an evidence prototype in the chunk embedding space, and the candidate most aligned with this prototype is selected by cosine similarity. For stricter deployment budgets, we further introduce an optional task-aware feature selection strategy to reduce the selector input dimension. To support supervised evaluation, we construct semantic chunk-correctness labels based on evidence sufficiency rather than answer-string containment. Experiments on TriviaQA, PopQA, and MS MARCO Passage Ranking show that the proposed selector consistently improves rank-1 evidence selection over mobile-applicable baselines by an average of 2.5 percentage points. These results suggest that using LLM-side query representations and MoE routing information and aligning them with retrieval-side candidate embeddings is an effective and parameter-efficient strategy for mobile-applicable RAG chunk selection.
\end{abstract}
\begin{graphicalabstract}

\includegraphics[width=\linewidth]{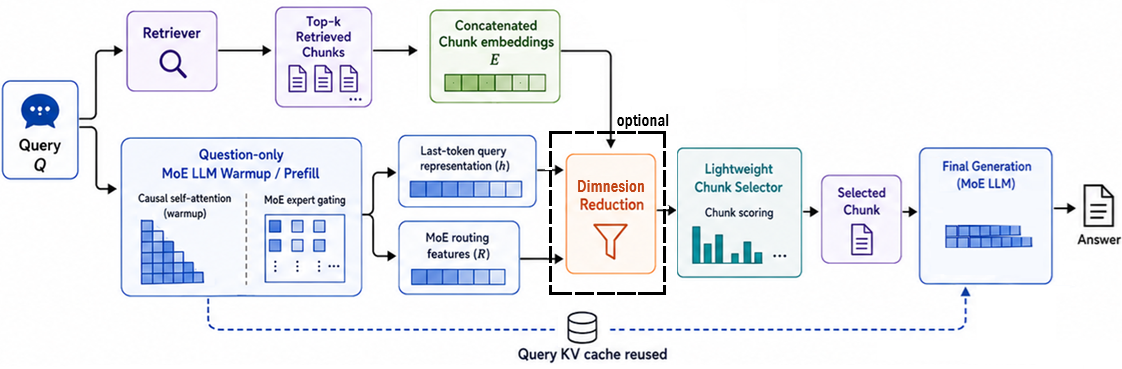}

\end{graphicalabstract}


\begin{highlights}
    \item Formulated mobile RAG chunk selection as a lightweight evidence-alignment problem.

    \item Proposed a chunk selection model using hidden states, MoE routing signals, and chunk embeddings.

    \item Introduced semantic chunk-correctness labels based on evidential sufficiency.

    \item Provided a budget-aware feature selection strategy for resource-constrained deployment settings.

    \item Achieved the best rank-1 chunk-selection accuracy across TriviaQA, PopQA, and MS MARCO.
\end{highlights}

%% Keywords

\end{frontmatter}



\section{Introduction}

Retrieval-augmented generation (RAG) has become a common framework for improving factual grounding by connecting large language models (LLMs) with external knowledge sources \cite{lewis2020rag}. It is especially useful in applications that require private data access, domain-specific knowledge, or timely information updates. Meanwhile, sparse mixture-of-experts (MoE) architectures have become increasingly prominent in recent LLM families \cite{jiang2024mixtral,deepseekai2024deepseekv3,yang2025qwen3}. There is also growing interest in deploying RAG systems on mobile and edge devices, where inference is constrained by computation, memory, latency, and energy budgets. In these settings, retrieved context increases prefill computation, memory use, and end-to-end latency, making context management an important issue for resource-constrained RAG \cite{kang2026pocketrag}.

A direct way to reduce the context cost is to restrict the generator to a single retrieved chunk. This aggressive context-reduction strategy can substantially reduce prompt length and token-level computation, but it turns chunk selection into a critical bottleneck. The top-ranked retrieved chunk is not necessarily the best evidence chunk, because dense retrieval optimizes semantic similarity rather than evidential sufficiency~\cite{karpukhin2020dpr}. Retrieved passages may be topically related to the query, mention the same entity, or share surface expressions without containing the fact or relation needed to answer the question. Thus, efficient mobile RAG requires a lightweight selector that identifies the most evidence-supporting chunk from the retrieved candidate pool instead of directly relying on retrieval rank.

Existing methods for reducing retrieved context can be broadly grouped into prompt compression, token- or sentence-level pruning, and chunk-level reranking. Prompt compression and rewrite-based methods reduce the final input length, but they usually require an additional LLM or compressor before generation such as LLMLingua~\cite{jiang2023llmlingua, pan2024llmlingua2} and RECOMP~\cite{xu2023recomp}. Token-level pruning methods such as Provence are more targeted to RAG context reduction, but still introduce an additional large reranker-pruner model~\cite{chirkova2025provence}. Cross-encoder rerankers can provide strong estimation, but they jointly encode each query-chunk pair therefore add inference cost for every candidate chunk~\cite{nogueira2019bert}. Recent rerankers such as BGE-reranker-v2-m3 contain hundreds of millions of parameters~\cite{bgererankerv2}, which makes them strong server-side baselines but less aligned with a strict mobile additional-module budget.
Under a constrained edge-device budget, the most practical alternatives are retrieval-order methods, lexical rerankers, classical learning-to-rank models, and compact neural interaction models. These methods avoid an additional large LLM or compressor, but their relevance scores are derived primarily from surface matching or retriever-side query--document interactions. They therefore estimate passage relevance independently of the generator that will later consume the selected evidence. This generator-independent scoring creates a mismatch: a passage may be semantically related to the query while still lacking the specific fact or relation required by the downstream generator.

We address this mismatch by reusing a generator-native representation of
the question. Unlike the query embedding produced by a separate retrieval
encoder, the final question hidden state is generated by the same LLM
that will consume the selected evidence. Recent work has shown that LLM
hidden states can be projected directly into retrieval embedding spaces,
indicating that they retain retrieval-relevant information
\cite{jiang2026native}. We therefore use the final-token hidden state to
represent the generator's semantic interpretation of the query.

For MoE language models, routing behavior provides an additional
structured representation of the query. Prior work has shown that MoE
routing weights can serve as embedding-like representations and provide
information complementary to hidden-state embeddings
\cite{li2025moeembedding,11517464, raza2025neuromoetransformerbasedmixtureofexpertsframework, shahcogmoe, Raza2026CVPR}. More broadly, analysis indicates that
expert use patterns encode structured information about 
processed input
\cite{muennighoff2024olmoe,bandarkar2026multilingual, fedus2022switch}. Motivated by these findings, we use the
final question hidden state and routing-weighted expert-response features
as complementary generator-side signals. The hidden state summarizes the
resulting query representation, while the routing-derived feature retains
structured information about expert utilization across layers.

These generator-side features characterize the information need encoded
by the language model, but they do not represent the differences among
the retrieved candidates. We therefore combine them with dense chunk
embeddings, which provide compact candidate-specific representations in
a shared retrieval space and use a multilayer perceptron (MLP) to
predict an evidence prototype in the same embedding space. Each
candidate is then scored by the cosine similarity between its embedding
and the predicted prototype. 
For more resource-constrained settings, we further introduce an optional
task-aware feature-selection module that retains the most informative
input dimensions before evidence-prototype prediction. In the RAG
pipeline, retrieval proceeds in parallel with the question-only language
model warmup, and the resulting query-side KV cache is reused during
subsequent generation. The remaining online selection cost is therefore
limited to the compact selector and cosine-similarity scoring.

Another challenge is the lack of reliable chunk-level supervision. Existing open-domain QA datasets provide final answers, but they do not directly specify which retrieved chunk should be retained before generation. A common lexical heuristic is to mark a chunk as correct if it contains the reference answer string, but this criterion is noisy for RAG chunk selection: a chunk may mention the answer string without providing the relation needed to answer the question, while another chunk may support the answer through paraphrase, aliasing, or indirect description. We therefore define chunk correctness by evidence sufficiency rather than answer-string containment, enabling supervised training and evaluation of pre-generation chunk selection.

Across TriviaQA, PopQA, and MS MARCO datasets, the proposed selector achieves the highest rank-1 chunk-selection accuracy among the evaluated lightweight baselines, with an average improvement of 2.49 percentage points over the strongest baseline. The feature-budget analysis further shows that reduced-input variants retain competitive performance under tighter selector budgets.

The contributions of this paper are as follows:
\begin{itemize}
    \item We formulate resource-constrained RAG chunk selection as a LLM-coupled evidence selection problem, where query-side features are reused from the language-model warmup stage.
    \item We propose a lightweight chunk selection model that combines question hidden states, routing derived expert signals, and retrieved chunk embeddings in a compact multilayer perceptron.
    \item We show that our proposed method outperforms current mobile-RAG methods across three different datasets using only a compact trainable selection module.
    \item We introduce a chunk correctness dataset for RAG chunk selection, which is designed to study efficient context reduction under a supervised setting.
\end{itemize}

\section{Related Work}
Retrieval-augmented generation (RAG) systems often add a post-retrieval stage to reduce input length, suppress retrieval noise, and lower generation latency. Existing approaches can be organized into four categories: (i) token-level pruning or compression, (ii) structured extraction and selection, (iii) rewrite-based condensation that summarizes or synthesizes across top retrieved contexts, (iv) lightweight lexical and neural reranking.

\subsection{Token-Level Context Compression}
Token-level methods shrink the generator input by deleting or compressing lexical units under a token budget. \textit{Selective Context} identifies redundancy via language-model-based token scoring and prunes low-utility regions, yielding substantial reductions in inference memory and latency while maintaining downstream quality \cite{li-etal-2023-compressing}. Prompt-compression approaches such as \textit{LLMLingua} perform coarse-to-fine token removal with a budget controller to preserve semantics under high compression ratios \cite{jiang-etal-2023-llmlingua}, and \textit{LLMLingua-2} further formulates compression as token classification trained via LLM distillation to improve faithfulness and efficiency \cite{pan2024llmlingua2}. Query-guided attention compression leverages question-to-context attention to filter tokens under explicit length constraints \cite{wang2024quito}. 
Although token-level compression reduces the generator’s prompt length, it often introduces non-trivial end-to-end overhead and can delete discourse-critical cues, making the resulting fragmented context harder for small on-device LLMs to use reliably.

\subsection{Structured Extraction and Selection}
Structure-preserving methods keep linguistic units intact while removing irrelevant content. At the sentence/span level, \textit{RECOMP} includes an extractive compressor that selects useful sentences and an abstractive compressor that synthesizes across documents; it also supports selective augmentation by outputting empty context when retrieval is unhelpful \cite{xu2023recomp}. \textit{FILCO} targets imperfect retrieval by learning to filter contexts using lexical and information-theoretic signals, improving robustness across knowledge-intensive tasks \cite{wang2023filco}. In deployed stacks, embedding-based optimizers similarly remove low-similarity sentences to reduce token usage \cite{llamaindexsentenceembeddingoptimizer}.
At the passage level, systems often select a subset of retrieved chunks and/or apply reranking. Modern RAG toolkits expose strong rerankers, frequently cross-encoders that jointly score each $(q, p_i)$ pair to reorder candidates beyond embedding similarity \cite{haystackrankers}. However, reranking introduces additional inference and memory overhead,
which can be costly for interactive on-device RAG.

\subsection{Rewrite-Based Condensation}
Rewrite-based condensing replaces retrieved text with a shorter query-aware representation. \textit{RECOMP} includes an abstractive compressor trained for end-task utility while keeping summaries concise \cite{xu2023recomp}. \textit{CompAct} actively condenses extensive retrieved documents via iterative refinement to retain multi-hop evidence under high compression rates \cite{yoon-etal-2024-compact}. Hierarchical summary substitution approaches such as \textit{RAPTOR} build multi-resolution summaries and retrieve from a tree of abstractions at inference time \cite{sarthi2024raptor}. Tooling-level implementations also expose LLM-based extractors that rewrite/extract relevant parts of each retrieved document \cite{langchainllmchainextractor}. While effective in reducing prompt length, rewrite-based methods typically require additional LLM or extra compressor before answering, significantly increasing on-device latency and energy.

\subsection{Lightweight Lexical and Neural Reranking}

In addition to context compression and extraction, another line of work reorders retrieved candidates using lexical, learning-to-rank, or compact neural ranking models which are most suitable to mobile-applicable chunk selection because they avoid an additional large LLM or heavy context-compression model at inference time. BM25 remains a strong lexical ranking baseline based on term-frequency and document-length normalization under the probabilistic relevance framework~\cite{robertson2009bm25}. Classical learning-to-rank methods such as RankSVM~\cite{joachims2002optimizing} and LambdaMART~\cite{burges2010ranknet} learn ranking functions from supervised relevance signals and are widely used as efficient reranking baselines.

Neural interaction models further improve ranking by modeling query-document matching patterns. DUET combines local exact-match signals with distributed semantic representations for web search ranking~\cite{mitra2017duet}. DRMM models ad-hoc retrieval at the query-term level using matching histograms, a feed-forward matching network, and term gating~\cite{guo2016drmm}. KNRM introduces kernel pooling over word-level similarity matrices to capture multi-level soft matching patterns in an end-to-end ranking model~\cite{xiong2017knrm}. These models provide compact neural alternatives to large cross-encoder rerankers, but their relevance estimates are still derived from query-document matching alone. They do not exploit the internal representations of the generator that will produce the final answer, such as hidden states or MoE routing behavior. 

\section{Methodology}
\label{sec:methodology}

% Our RAG method is coupled with the LLM inference pipeline. Given a user query, the retriever and the language-model warmup stage are launched in parallel. The retriever returns the top-K candidate chunks, while the LLM performs a question-only forward pass to prefill the query tokens. This warmup pass is already required by the LLM inference, and the resulting query-side KV cache can be reused after chunk selection. Therefore, the MoE routing features and query hidden state used by the selector are extracted from computation shared with the subsequent generation stage.

Our RAG method is coupled with the LLM inference pipeline. Given a user query, the retriever and the language-model warmup stage are launched in parallel. The retriever returns the top-K candidate chunks, while the LLM performs a question-only forward pass to prefill the query tokens. This warmup pass is already required by the LLM inference, and the resulting query-side KV cache can be reused after chunk selection. For MoE models, the routing patterns generated during this pass provide a layer--expert representation of how computation is allocated for the query. Prior work has shown that such routing representations contain information complementary to hidden states and can function as embedding-like features~\cite{li2025moeembedding}. Therefore, the MoE routing features and query hidden state used by the selector are extracted from computation shared with the subsequent generation stage.

\begin{figure}[t]
\centering
\centering
\includegraphics[width=\textwidth]{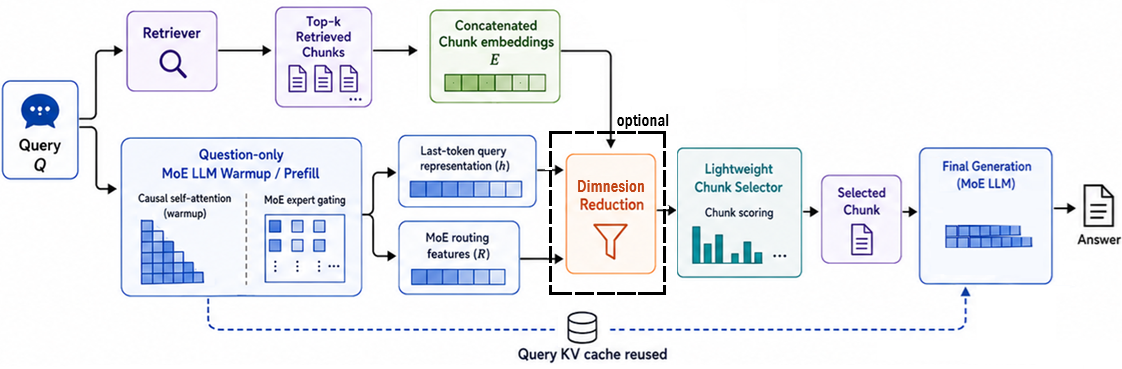}
\caption{Model overall architecture}
\label{fig:rag_arch}
\end{figure}

After the candidate chunks and warmup-stage query features become available, the selector estimates which retrieved chunk is most likely to provide useful evidence. The selector combines three feature sources: the final-token hidden state of the query, the MoE routing weights, and the embeddings of the retrieved candidate chunks. It then predicts an evidence vector in the chunk embedding space, and the final chunk is selected by cosine similarity between this predicted vector and each candidate chunk embedding. For more resource-constrained settings, we further introduce an optional budget-aware feature selection strategy that applies a task-aware feature mask to the extracted representation before the chunk selector, thereby reducing the selector input dimension. The overall pipeline is shown in Figure~\ref{fig:rag_arch}.

% We propose a lightweight chunk selection framework for retrieval-augmented generation that identifies the most useful retrieved chunk while minimizing computational cost. The framework is architected to bypass the fine-grained query-chunk interactions typical of conventional reranking approaches. The proposed methodology consists of three core components. First, we construct a structured input representation that combines question-only inference signals from a Mixture-of-Experts language model with dense embeddings of the top-K retrieved candidate chunks. Second, we perform importance-based dimension reduction to retain the most informative features and suppress redundant dimensions. Third, we develop a compact multilayer perceptron that maps the reduced representation into the chunk embedding space and selects the chunk most likely to support downstream answer generation.

\subsection{Input Representation}
\label{subsec:input-rep}

To utilize information from both the query and the retrieved candidates, we construct the input representation from three complementary feature sources: the question hidden state, the MoE routing-derived expert response, and the dense embeddings of the retrieved chunks. The question-side features describe how the language model internally represents the query, while the chunk-side features preserve the semantic structure of the candidate evidence produced by the retriever.

For each query $q$, the input vector is defined as
\begin{equation}
x = \big[\, h \;\Vert\; r \;\Vert\; E \,\big] \in \mathbb{R}^{4992},
\label{eq:input-vector}
\end{equation}
where $h \in \mathbb{R}^{2048}$ is the question hidden state, $r \in \mathbb{R}^{1024}$ is the routing-score feature, and $E \in \mathbb{R}^{1920}$ is the flattened representation of the top-K retrieved chunk embeddings. In this work, we use $K=5$, and each retrieved chunk is represented by a $384$-dimensional dense embedding.

The hidden state $h$ is extracted from the final-layer last-token representation of the MoE language model under question-only input. This representation provides a compact summary of the query semantics before any retrieved context is added. Because it is obtained from a single question-only forward pass, it avoids the repeated inference required by cross-encoder reranking, where every query--chunk pair must be processed separately.

In addition to the hidden state, we use the routing behavior of the MoE model as a complementary query-side signal. The hidden state captures the final semantic representation of the query, whereas the routing pattern reflects how the model distributes the query across different experts during inference. However, the raw router probability only represents the model's allocation preference before expert transformation and does not directly measure the realized contribution of an expert to the resulting representation. An expert may receive a high routing probability but produce an output with a small magnitude, while another expert with a lower routing probability may exert a stronger representation-level effect through a larger response. We therefore combine the routing probability with the magnitude of the corresponding expert output.

For each layer--expert pair, we compute a routing-weighted expert-response score by averaging over all question tokens:
\begin{equation}
\mathrm{RS}_{l,j}
=
\frac{1}{T}
\sum_{t=1}^{T}
p_{l,j}^{(t)}
\left\|
g_{l,j}\!\left(u_l^{(t)}\right)
\right\|_2,
\label{eq:rweon}
\end{equation}
where $T$ is the number of question tokens, $p_{l,j}^{(t)}$ is the router probability assigned to expert $j$ at layer $l$ for token $t$, $u_l^{(t)}$ denotes the input representation to the MoE block, and $g_{l,j}(\cdot)$ denotes the output of expert $j$. The product in Eq.~\ref{eq:rweon} provides a closer approximation of the realized expert contribution than the router probability alone.

We use the routing probabilities over all $64$ experts rather than retaining only the top-$8$ routing values. The resulting scores from all $16$ layers and $64$ experts per layer are flattened to form the routing feature $r$.

For the candidate side, each retrieved chunk $c_k$ is represented by an $\ell_2$-normalized dense embedding $e_k \in \mathbb{R}^{384}$, where $\|e_k\|_2 = 1$. The top-K chunk embeddings are stacked and flattened as
\begin{equation}
E =
\operatorname{vec}
\left(
\begin{bmatrix}
e_0^\top;
e_1^\top;
\cdots;
e_{K-1}^\top
\end{bmatrix}
\right)
\in \mathbb{R}^{384K}.
\label{eq:chunk-embedding-block}
\end{equation}
With $K=5$, this gives $E \in \mathbb{R}^{1920}$.

Overall, $h$ and $r$ provide query-side information derived from a single language-model pass, while $E$ provides the candidate-side retrieval information. The selector then learns to combine these features to estimate which retrieved chunk is most useful for downstream generation, without explicitly modeling token-level interactions between the query and every candidate chunk.

\subsection{Budget-Aware Dimension Reduction}
\label{subsec:dim-reduction}

\begin{figure}[t]
\centering
\centering
\includegraphics[width=0.6\textwidth]{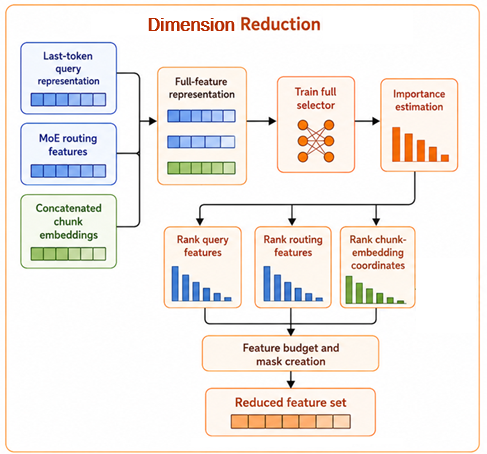}
\caption{Budget-Aware Optional Dimension Reduction Module}
\label{fig:rag_arch_dimension_reduction}
\end{figure}

The full input representation is high-dimensional because it combines the question hidden state, routing-score feature, and retrieved-chunk embeddings. Although these features provide complementary information for chunk selection, directly using all dimensions can be inefficient and may complicate learning. However, mobile and edge deployment may impose stricter constraints on selector size, memory bandwidth, and input dimensionality. We therefore perform an optional dimension reduction to retain the dimensions most relevant to the supervised chunk-selection objective, as shown in Figure \ref{fig:rag_arch_dimension_reduction}.

Instead of applying an unsupervised compression method, we estimate feature importance from the trained selector. This is because variance-based compression does not necessarily preserve dimensions that are useful for evidence selection. A feature dimension with small variance may still be discriminative if the selector consistently relies on it to distinguish useful chunks from irrelevant ones. We therefore use gradient-based sensitivity as a task-aware importance measure.

Specifically, we first train a full-dimensional selector using the original input $x \in \mathbb{R}^{4992}$. After training, the selector is frozen, and the importance of each input dimension is measured by the accumulated absolute gradient of the loss with respect to that dimension:
\begin{equation}
I_d =
\sum_{t \in \mathcal{T}}
\left|
\frac{\partial \mathcal{L}_t}{\partial x_d}
\right|,
\label{eq:grad-importance}
\end{equation}
where $\mathcal{T}$ denotes the training set and $\mathcal{L}_t$ is the loss for training sample $t$. A larger $I_d$ indicates that changes along dimension $d$ have a stronger effect on the selection loss, and therefore that the dimension is more important for the trained selector.

For the question hidden-state and routing-score blocks, dimensions are ranked directly according to $I_d$. For the chunk-embedding block, we apply an additional constraint. Since all retrieved chunks are represented in the same embedding space, the same embedding coordinate should be kept or removed consistently across all chunk positions. Otherwise, different candidates would be compared using different subspaces, which could introduce an artificial position-dependent bias.

Therefore, for each chunk-embedding coordinate $p$, we aggregate its importance across the $K$ retrieved chunks:
\begin{equation}
I^{\mathrm{chunk}}_p =
\frac{1}{K}
\sum_{k=0}^{K-1}
I_{E_{k,p}},
\label{eq:chunk-avg-importance}
\end{equation}
where $I_{E_{k,p}}$ denotes the importance of coordinate $p$ in the embedding of the $k$-th retrieved chunk. If coordinate $p$ is selected, it is retained for all retrieved chunks. This preserves the shared embedding geometry used for candidate comparison.

Given a target retention ratio $X\%$, we define the feature budget as $\left\lfloor 4992 \cdot X/100 \right\rfloor$ with respect to the original input size. Selecting one hidden-state or routing-score dimension consumes one unit of this budget, while selecting one chunk-embedding coordinate consumes $K$ units because the coordinate is retained for all $K$ chunks. The selected dimensions form a feature mask, which is applied to the original input to produce a reduced representation $x' \in \mathbb{R}^{D}$.

After the mask is obtained, we retrain the MLP selector using the reduced representation $x'$ and directly apply the same mask during inference. This two-stage process allows the full-dimensional model to first identify task-relevant dimensions, and then replaces it with a smaller selector trained only on the retained features.

\subsection{Chunk Selection}
\label{subsec:selection-net}

\begin{figure}[t]
\centering
\centering
\includegraphics[width=0.6\textwidth]{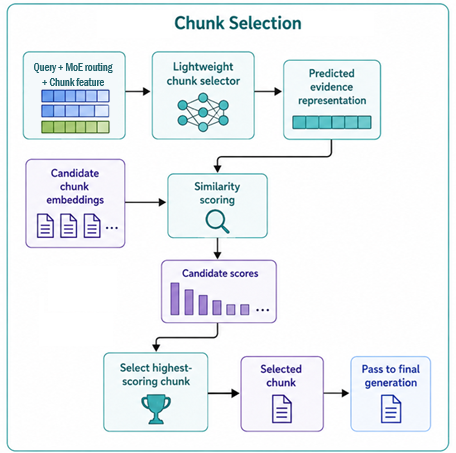}
\caption{Chunk Selection Module}
\label{fig:rag_arch_chunk_select}
\end{figure}

Given the reduced representation $x'$, the selector predicts an evidence prototype in the same embedding space as the retrieved chunks:
\begin{equation}
\hat{y} = f_\theta(x') \in \mathbb{R}^{384},
\label{eq:selector-output}
\end{equation}
where $f_\theta$ is a compact multilayer perceptron. Rather than directly predicting a retrieval position, the selector estimates the location of supporting evidence in the retriever embedding space. This formulation allows the predicted prototype to be compared with all candidate chunks using the semantic geometry of the retriever. Although the ordered candidate set may implicitly preserve retrieval-rank information, retrieval rank is not used as the supervision target.

Before candidate scoring, the predicted prototype is $\ell_2$-normalized as $\tilde{y}=\hat{y}/\|\hat{y}\|_2$. Since each retrieved chunk embedding $e_k$ is also normalized, its alignment with the predicted prototype is computed by cosine similarity:
\begin{equation}
s_k = \mathrm{sim}(\tilde{y}, e_k)
    = \tilde{y}^{\top} e_k.
\label{eq:chunk-similarity}
\end{equation}
The resulting scores quantify the relative alignment of the retrieved candidates with the predicted evidence prototype and are used to prioritize the candidate set before generation.

\paragraph{Multi-target training objective}
A query may be supported by more than one retrieved chunk. Therefore, using a single-label objective can incorrectly penalize the selector when it assigns high similarity to another chunk that also contains valid evidence. To better match this property of RAG evidence selection, we train the selector with a multi-target contrastive loss.

For training sample $i$, let $\mathcal{P}_i$ denote the set of positive chunk embeddings that contain valid answer evidence, and let $\mathcal{N}_i$ denote the remaining retrieved chunk embeddings. The loss is defined as
\begin{equation}
\mathcal{L}
=
-\frac{1}{B}
\sum_{i=1}^{B}
\log
\frac{
\sum\limits_{p \in \mathcal{P}_i}
\exp\!\big(\mathrm{sim}(\tilde{y}_i,p)/\tau\big)
}{
\sum\limits_{c \in \mathcal{P}_i \cup \mathcal{N}_i}
\exp\!\big(\mathrm{sim}(\tilde{y}_i,c)/\tau\big)
},
\label{eq:multi-infonce}
\end{equation}
where $\tau$ is the temperature parameter and $B$ is the batch size. This objective pulls the predicted vector toward any answer-supporting chunk while reducing its similarity to retrieved chunks that do not contain useful evidence.

Training samples with no positive chunks are excluded because they do not provide a supervised target for selection. Samples with no negative chunks are also excluded because they do not define a meaningful contrastive comparison among retrieved candidates.

At inference time, the same feature mask is applied to the input representation to obtain $x'$. The selector then predicts $\tilde{y}$, and each retrieved chunk is scored by cosine similarity. The final selected chunk is
\begin{equation}
k^*
=
\arg\max_{k \in \{0,\dots,K-1\}}
\mathrm{sim}(\tilde{y}, e_k).
\label{eq:inference-selection}
\end{equation}
The selected chunk $c_{k^*}$ is paired with the original query for downstream generation. In this way, the selector filters retrieved context without applying token-level query--chunk interaction to every candidate.

\section{Dataset and Chunk-Level Ground-Truth Construction}
\label{sec:dataset}

Our task requires supervision at the retrieved-chunk level rather than only at the final-answer level. Existing open-domain question answering datasets provide question--answer pairs, but they do not directly indicate which retrieved candidates contain sufficient evidence for answering the question. A common alternative is to derive chunk labels from answer-string containment. However, lexical containment is only an approximate indicator of evidential support and can produce both false-positive and false-negative labels.

To obtain supervision that is better aligned with chunk selection, we construct semantic support labels for the retrieved candidates. Section \ref{sec:gt_construction} describes the annotation criterion and labeling procedure, while Section \ref{sec:dataset_structure} presents the resulting instance structure and its instantiation on TriviaQA, PopQA, and MS MARCO Passage Ranking.

\subsection{Chunk-Level Ground-Truth Construction}
\label{sec:gt_construction}

\begin{figure}[t]
\centering
\centering
\includegraphics[width=\textwidth]{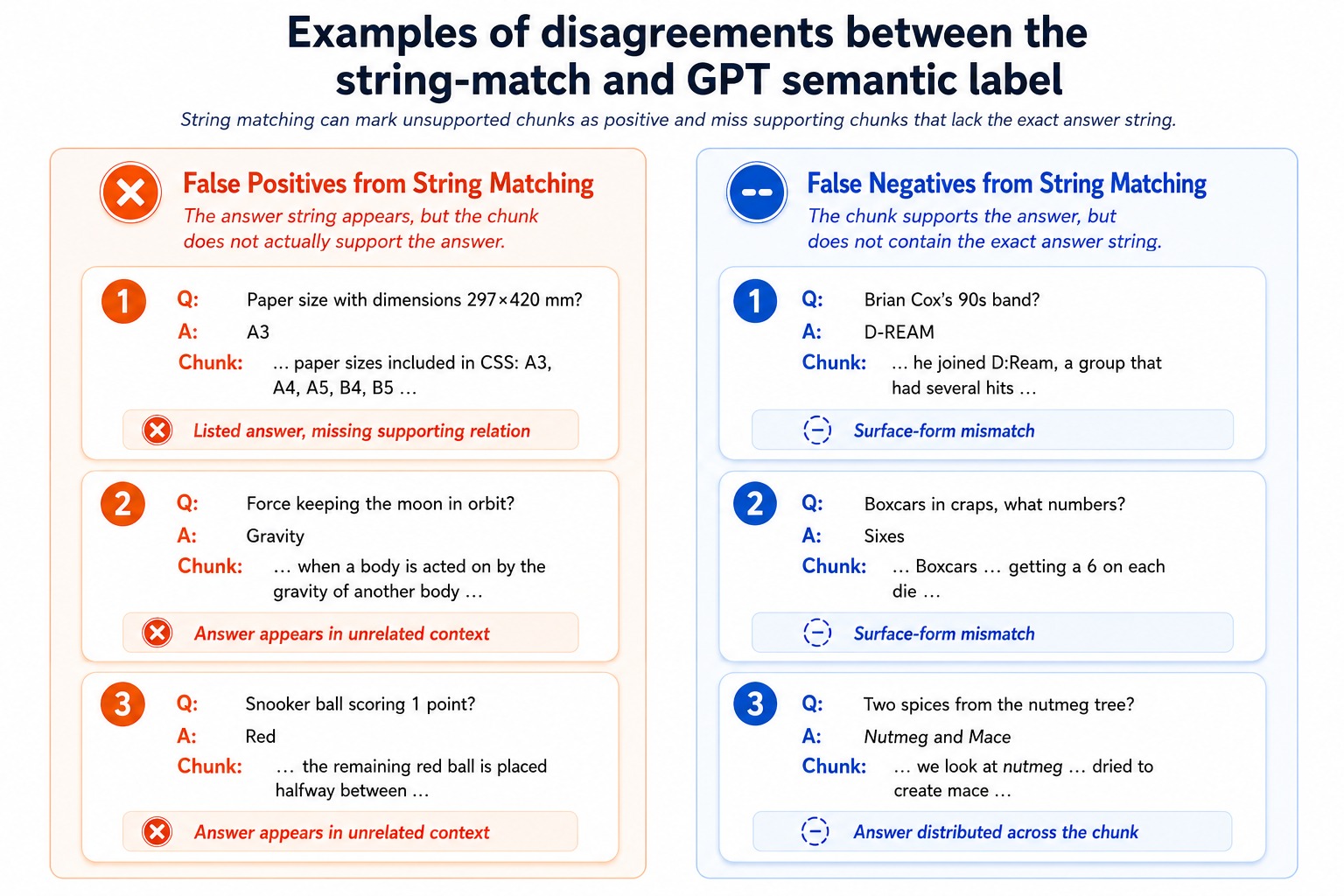}
\caption{Examples of disagreements between the string-match \texttt{contains\_answer} label and the GPT-5.1 semantic support label on TriviaQA. }
\label{fig:relabel_example}
\end{figure}

Answer-string containment does not always correspond to evidential support, as illustrated in Figure~\ref{fig:relabel_example}. Two types of disagreement are particularly relevant. First, a retrieved chunk may contain the reference answer string without expressing the relation needed to answer the question. For example, the answer may occur in a list, a biographical note, a disambiguation passage, or an otherwise unrelated discussion. Such cases are false positives under lexical labeling. Second, a chunk may provide sufficient supporting evidence without containing the exact reference string, for example through paraphrases, aliases, or distributed descriptions. These cases are false negatives under lexical labeling.

To address these limitations, we construct semantic support labels for the retrieved candidate chunks. Each question--chunk pair is evaluated using GPT-5.1 as an automated annotator. The annotator is asked to determine whether the chunk provides sufficient evidence to answer the question using only the supplied context. This procedure defines chunk correctness by evidential sufficiency rather than exact answer-string occurrence.

\noindent\textbf{Prompt template.}
\begin{verbatim}
Can you find the evidence in the context to answer the question?
Answer only with Yes/No. Don't use any external tool. Don't search
on web. Only based on context, don't infer.
Question: {question}
Context: {context}
\end{verbatim}

A chunk is labeled positive if it contains enough information to answer the given question. A chunk is labeled negative if it does not provide relevant evidence, even when the answer string appears in the chunk. In a preliminary comparison, the original answer-containment labels agree with the semantic support labels for 78\% of the evaluated chunks, indicating that lexical matching is useful but introduces substantial label noise. The semantic verification process therefore serves two purposes: it recovers supporting chunks missed by lexical matching, and it removes spurious positives caused by accidental answer-string overlap. The resulting semantic support labels serve as the ground truth for training and evaluating the chunk selector.

\subsection{Dataset Structure and Instantiation}
\label{sec:dataset_structure}

Each dataset instance is constructed from a question and reference answer drawn from the corresponding question answering benchmark, together with five candidate chunks retrieved from a separate passage corpus. Candidate chunks are annotated with semantic support labels indicating whether they provide sufficient evidence for the reference answer. An instance is considered answerable if at least one retrieved chunk receives a positive semantic support label. Otherwise, it is treated as unanswerable within the retrieved set.

\paragraph{TriviaQA}
TriviaQA is an open-domain question answering benchmark constructed from Wikipedia articles. Questions typically focus on encyclopedic facts, entities, events, and concepts that can often be answered using a single supporting passage. For each TriviaQA question, candidate chunks are retrieved from a static index of DPR-style 100-word Wikipedia passages, \texttt{psgs\_w100}, containing approximately 21 million passages. Since the underlying retrieval corpus is relatively clean and well structured, TriviaQA provides a representative benchmark for evaluating evidence selection in a controlled retrieval environment.

\paragraph{PopQA}
PopQA is an entity-focused question answering benchmark built from knowledge graph relations. The dataset covers facts associated with entities spanning a wide range of popularity levels, including both frequently mentioned and long-tail entities. Compared with traditional Wikipedia-based benchmarks, PopQA places greater emphasis on retrieving evidence for less common facts, making it useful for evaluating robustness across different levels of knowledge frequency. For each PopQA question, candidate chunks are retrieved from the same static index of DPR-style 100-word Wikipedia passages used for TriviaQA.

\paragraph{MS MARCO Passage Ranking}
MS MARCO is a large-scale retrieval benchmark derived from real-world search queries and web documents. In contrast to the Wikipedia-based retrieval corpus used for TriviaQA and PopQA, its passages originate from diverse web sources and exhibit substantial variation in writing style, document quality, and relevance. We use the MS MARCO Passage Ranking task (MS MARCO PR), with candidate chunks retrieved from the official MS MARCO passage collection, containing approximately 8.8 million passages. Retrieved passages frequently contain partial answers, redundant information, or references to answer entities without providing sufficient evidence to answer the question. These characteristics make evidence identification substantially more challenging, resulting in a more difficult chunk selection task than in TriviaQA or PopQA.

For all datasets, semantic support labels are constructed using the procedure described in Section~\ref{sec:gt_construction}. Given a question and its retrieved candidate chunks, chunk selection is formulated as a supervised prediction problem in which the model identifies the candidate most likely to provide sufficient evidence for answering the question. Together, TriviaQA, PopQA, and MS MARCO PR allow evaluation under increasingly challenging retrieval conditions, ranging from relatively clean encyclopedic passages to long-tail factual knowledge and diverse web documents.

\section{Experimental Setup}
\label{sec:experimental-setup}

\subsection{Retrieval Configuration}
\label{subsec:retrieval-config}

Passages and queries are encoded using the \texttt{all-MiniLM-L6-v2} Sentence-Transformer model \cite{reimers2019sentencebert,minilm_model} with $\ell_2$-normalized embeddings. For scalable retrieval, we construct a FAISS IVF-PQ index \cite{douze2024faiss} configured with 8192 clusters, 64 subquantizers, and 8 bits per subquantizer. The quantizer is trained on a uniform sample of 500,000 embeddings per corpus. During query execution, we search 32 clusters and retain the top five passages. All experiments share this uniform retrieval and candidate-generation protocol across benchmarks.

\subsection{Feature Extraction and Selector Implementation}
\label{subsec:implementation}

Query-side features are extracted using the frozen \texttt{OLMoE-1B\allowbreak-7B\allowbreak-0924\allowbreak-Instruct}
 model, which features 16 layers, 64 experts per layer. Forward hooks on all MoE layers collect the required hidden-state and routing information without updating the backbone parameters. Feature extraction runs on a 16GB NVIDIA Tesla V100 GPU using 4-bit NF4 and double quantization with FP16 computation. Questions are formatted using the official instruction template truncated to 512 tokens, strictly excluding retrieved chunks.

Selector inputs undergo standard scaling. The selector is a 7-million-parameter multilayer perceptron comprising three 1024-width hidden layers, each equipped with batch normalization, ReLU activation, and a dropout rate of 0.05. The final output vector is $\ell_2$-normalized prior to candidate scoring.

Optimization utilizes AdamW with a learning rate of $8 \times 10^{-4}$, weight decay of $3 \times 10^{-5}$, and a batch size of 256. We apply Kaiming normal initialization, gradient clipping at a norm of one, and a OneCycleLR schedule with cosine annealing. Models are trained for up to 150 epochs with a 20-epoch early stopping patience based on validation loss. Instances lacking either positive or negative candidates are filtered out during optimization.

\section{Results}
\label{sec:results}
\subsection{Comparison with SOTA methods}

\begin{table}[t]
\centering
\caption{Rank-1 chunk-selection accuracy (\%) under a mobile-applicable selection budget. The parameter column reports additional trainable parameters introduced by the selection module. The best result on each dataset is shown in bold.}
\label{tab:main-results}
\resizebox{\linewidth}{!}{
\begin{tabular}{lrrrr}
\toprule
\textbf{Method} & \textbf{Params} & \textbf{TriviaQA} & \textbf{PopQA} & \textbf{MS MARCO PR} \\
\midrule
RAG baseline & -- & 44.96 & 61.74 & 65.82 \\
BM25 (lexical) \cite{robertson2009bm25} & -- & 45.01 & 62.80 & 66.14 \\
RankSVM \cite{joachims2002optimizing} & -- & 46.13 & 64.46 & 67.49 \\
LambdaMart \cite{burges2010ranknet} & -- & 46.00 & 64.41 & 66.76 \\
Local-DUET \cite{mitra2017duet}& 49M & 46.28 & 63.33 & 66.97 \\
DRMM \cite{guo2016drmm}& 48M & 46.53 & 65.72 & 68.13 \\
KNRM \cite{xiong2017knrm}& 48M & 47.04 & 66.76 & 69.58 \\
\textbf{Ours} & 7M & \textbf{49.56} & \textbf{70.87} & \textbf{70.43} \\
\bottomrule
\end{tabular}
}

\end{table}

Table~\ref{tab:main-results} compares the rank-1 chunk-selection accuracy of our method with seven representative mobile-RAG baselines on three datasets. The parameter column reports the number of additional trainable parameters introduced by the reranking or selection module.

Our method achieves the best performance on all three datasets, demonstrating that the proposed lightweight selector can consistently improve the quality of the top-ranked evidence chunk. Compared with the retrieval-only RAG baseline, our method improves rank-1 accuracy by 4.60 points on TriviaQA, 9.13 points on PopQA, and 4.61 points on MS MARCO PR. These gains indicate that the first retrieved chunk is often not the most reliable evidence for generation, and that an explicit chunk-selection module is necessary even when the retriever already returns relevant candidates.

Compared with lexical and classical learning-to-rank methods, our method shows a clear advantage. BM25 only brings marginal improvements over the RAG baseline, suggesting that lexical matching alone cannot reliably distinguish truly supportive chunks from topically related but insufficient ones. RankSVM and LambdaMart further improve the ranking by using handcrafted ranking features, but their gains remain limited because they do not directly model fine-grained semantic alignment between the query and the candidate evidence. In contrast, our method learns a compact neural selection function that better captures evidence-level relevance, leading to stronger rank-1 selection accuracy across all datasets.

Among prior neural baselines, our method still outperforms KNRM by 2.52, 4.11, and 0.85 points on TriviaQA, PopQA, and MS MARCO PR, with an average gain of 2.5 points, while using only 7M additional parameters. This result is important because KNRM and DRMM rely on substantially larger neural interaction modules under our parameter accounting, whereas our model achieves better accuracy with a much smaller selection network. Therefore, the improvement is not only in absolute accuracy, but also in the accuracy--efficiency trade-off, which is crucial for mobile-device RAG.

The largest gain appears on PopQA. This is reasonable because PopQA contains many entity-centric factual questions, where multiple retrieved chunks may share similar entities or surface-level keywords, but only one chunk provides the direct factual support needed for answering. Our method is better able to identify this evidence-bearing chunk, which explains the large improvement over both lexical and neural baselines. On TriviaQA, the absolute margin is smaller because all methods perform within a narrow range, suggesting that the main bottleneck may come from candidate retrieval quality or ambiguous evidence distribution rather than reranking capacity alone. On MS MARCO PR, our method also obtains the best result, showing that the proposed selector generalizes beyond open-domain factual question-and-answer to passage-ranking-style retrieval scenarios.

\subsection{Ablation Study}
\label{sec:ablation}

\begin{table}[t]
\centering
\caption{Ablation study of the proposed chunk selector. HS denotes hidden-state features, RW denotes routing-weight features, and Chunk denotes retrieved chunk embeddings. The best result on each dataset is shown in bold.}
\label{tab:ablation}
\resizebox{\linewidth}{!}{
\begin{tabular}{lccc}
\toprule
\textbf{Model} & \textbf{TriviaQA} & \textbf{PopQA} & \textbf{MS MARCO PR} \\
\midrule
Rank baseline & 44.96 & 61.74 & 65.82 \\
HS & 46.24 & 64.85 & 66.45 \\
RW & 44.98 & 62.50 & 65.93 \\
Chunk & 47.31 & 67.92 & 69.28 \\
HS + RW & 44.95 & 63.10 & 66.18 \\
HS + Chunk & 49.03 & 68.55 & 69.87 \\
RW + Chunk & 48.27 & 69.41 & 70.05 \\
HS + RW + Chunk (ours) & \textbf{49.56} & \textbf{70.87} & \textbf{70.43} \\
\bottomrule
\end{tabular}
}
\end{table}

Table~\ref{tab:ablation} presents the ablation results of the proposed chunk selector on TriviaQA, PopQA, and MS MARCO PR. Starting from the rank baseline, we progressively incorporate hidden-state features (HS), routing-weight features (RW), and retrieved chunk embeddings (Chunk) to evaluate the contribution of each component.

The results show that retrieved chunk embeddings are the most important source of performance improvement. Using chunk embeddings alone increases rank-1 accuracy from 44.96 to 47.31 on TriviaQA, from 61.74 to 67.92 on PopQA, and from 65.82 to 69.28 on MS MARCO PR. These gains are substantially larger than those obtained by HS or RW individually, indicating that semantic information encoded in the retrieved chunks provides the strongest signal for identifying answer-supporting evidence.

The contributions of HS and RW alone are more modest. HS consistently improves performance across all datasets, suggesting that the language model hidden state captures useful information about question semantics. In contrast, RW alone provides only limited gains, indicating that routing patterns are not sufficiently informative when used in isolation. Nevertheless, routing information appears to contain complementary signals that become useful when combined with stronger semantic features.

The pairwise combinations further reveal interactions among the three feature sources. Both HS + Chunk and RW + Chunk outperform Chunk alone on all datasets, demonstrating that hidden-state and routing information provide additional evidence beyond the retrieved chunk content itself. In particular, RW + Chunk achieves 70.05 on MS MARCO PR, while HS + Chunk reaches 49.03 on TriviaQA, showing that the usefulness of complementary signals varies across datasets.

The complete model achieves the best performance on all three benchmarks, reaching 49.56 on TriviaQA, 70.87 on PopQA, and 70.43 on MS MARCO PR. Compared with the rank baseline, the full model improves rank-1 chunk-selection accuracy by 4.60 points, 9.13 points, and 4.61 points on the three datasets, respectively. The largest improvement is observed on PopQA, where multiple retrieved chunks frequently contain similar entity mentions while only a subset provides sufficient evidence for answering the question. In such cases, combining semantic chunk representations with question-aware hidden states and routing signals enables more accurate evidence identification.

Overall, the ablation study demonstrates that retrieved chunk embeddings provide the primary source of selection accuracy, while hidden-state and routing-weight features contribute complementary information. The combination of all three components yields the most robust performance across diverse retrieval settings, validating the design of the proposed chunk-selection framework.

\subsection{Feature Budget Analysis}

\begin{table}[t]
\centering
\caption{Feature-budget analysis of the proposed selector. The full-feature model achieves the best accuracy, while reduced-feature models provide smaller operating points under stricter deployment budgets.}
\label{tab:feature-budget}
\resizebox{\linewidth}{!}{
\begin{tabular}{cccccc}
\toprule
\textbf{Feature Budget} & \textbf{\# Dims} & \textbf{TriviaQA} & \textbf{PopQA} & \textbf{MS MARCO PR} & \textbf{Avg.} \\
\midrule
10\%  & 499  & 45.83 & 66.19 & 67.42 & 59.81 \\
20\%  & 998  & 47.01 & 68.03 & 68.67 & 61.24 \\
30\%  & 1498 & 47.34 & 68.75 & 68.90 & 61.66 \\
40\%  & 1997 & 48.12 & 69.31 & 69.36 & 62.26 \\
50\%  & 2496 & 48.29 & 69.20 & 69.58 & 62.36 \\
60\%  & 2995 & 48.61 & 69.94 & 69.79 & 62.78 \\
70\%  & 3494 & 48.47 & 70.28 & 70.05 & 62.93 \\
80\%  & 3994 & 48.98 & 70.42 & 70.13 & 63.18 \\
90\%  & 4493 & 49.22 & 70.65 & 70.31 & 63.39 \\
100\% & 4992 & \textbf{49.56} & \textbf{70.87} & \textbf{70.43} & \textbf{63.62} \\
\bottomrule
\end{tabular}}
\end{table}

Table~\ref{tab:feature-budget} reports the performance of the selector under different retained-feature budgets. The full-feature model achieves the best accuracy on all three datasets, reaching 49.56 on TriviaQA, 70.87 on PopQA, and 70.43 on MS MARCO PR. This result confirms that the hidden-state, routing-derived, and chunk-embedding features provide complementary selection signals when the deployment budget allows.

The reduced-feature models show a clear accuracy--efficiency trade-off. Even with only 10\% of the original features, the selector improves over the retrieval-only rank baseline on all three datasets, increasing accuracy by 0.87 points on TriviaQA, 4.45 points on PopQA, and 1.60 points on MS MARCO PR. This indicates that the gradient-based feature budget retains a compact subset of informative dimensions rather than discarding features uniformly. The 20\% feature-budget model uses only 998 input dimensions and achieves 61.24 average
accuracy, which is better than the strongest compact neural baseline KNRM at 61.13 average accuracy. 

As the feature budget increases, performance improves smoothly and then gradually saturates. The 70\% feature-budget model reaches 62.93 average accuracy, and the 90\% model reaches 63.39, close to the full-feature result of 63.62. Therefore, feature selection provides a controllable operating curve: the full selector gives the best accuracy, while reduced selectors offer competitive alternatives when input dimensionality, parameter count, or memory bandwidth is more constrained.

% \section{Conclusion}
% \label{sec:conclusion}

% We presented a lightweight chunk-selection framework for retrieval-augmented generation. The method reuses question-only MoE representations and combines them with top-K chunk embeddings to predict an evidence vector, selecting the final chunk by cosine similarity. This avoids expensive pairwise query--chunk cross-encoding while allowing the selector to exploit model-side signals that are not available to lexical or kernel-pooling rankers.
% Across three datasets, Ours obtains the best rank-1 chunk-selection accuracy among all evaluated methods. It outperforms the strongest prior baseline by 2.5 points on average across three different datasets. These results suggest that a moderate-size selector under a limited parameter budget can provide a better accuracy--efficiency trade-off than both zero-parameter retrieval heuristics and very small neural IR rerankers.

\section{Conclusion}

This paper presented a lightweight chunk-selection framework for retrieval-augmented generation. The proposed method combines the question hidden state, routing-weighted MoE features, and retrieved chunk embeddings to select the correct chunk for LLM inference. To provide supervision for this task, we constructed semantic support labels based on evidential sufficiency rather than lexical answer containment. Across TriviaQA, PopQA, and MS MARCO Passage Ranking, the proposed selector achieved the highest rank-1 chunk-selection accuracy among the evaluated lightweight baselines, improving over the strongest baseline on each dataset by an average of 2.49 percentage points. 

The optional task-aware feature-selection procedure provides additional operating points under reduced selector-input budgets. Overall, the results show that generator-side query representations can be combined with retrieval-side candidate geometry to support compact, evidence-oriented chunk selection.

 \bibliographystyle{elsarticle-num} 
 \bibliography{custom}




% %% For numbered reference style

%   \textit{\LaTeX: a document preparation system},
%   Addison Wesley, Massachusetts,

% \end{thebibliography}
\end{document}